\pdfoutput=1

\documentclass[11pt]{article}

\usepackage{emnlp2021}

\usepackage{times}
\usepackage{latexsym}

\usepackage[T1]{fontenc}

\usepackage[utf8]{inputenc}

\usepackage{microtype}

\usepackage{color}
\usepackage{amsmath}

\usepackage{graphicx}
\usepackage{natbib}
\usepackage{caption}
\usepackage{amsfonts}
\usepackage{amssymb}
\usepackage{multirow}
\usepackage{adjustbox}
\usepackage[english]{babel} 
\usepackage{arydshln}
\usepackage[ruled,linesnumbered]{algorithm2e}
\usepackage{bbding}
\usepackage{CJKutf8}

\SetKwInOut{Parameter}{Parameters}

\title{\textit{Don't be Contradicted with Anything!} \\
  CI-ToD: Towards Benchmarking Consistency for Task-oriented \\ Dialogue System
}

\author{Libo Qin, Tianbao Xie, Shijue Huang, Qiguang Chen, Xiao Xu, Wanxiang Che\thanks{\ \  Email corresponding.} \\
 Research Center for Social Computing and Information Retrieval \\
 Harbin Institute of Technology, China \\
 {\tt \{lbqin,tianbaoxie,sjhuang,qgchen,xxu,car\}@ir.hit.edu.cn} 
}

\begin{document}
\maketitle
\begin{abstract}
\textit{Consistency Identification} has obtained remarkable success on open-domain dialogue, which can be used for preventing inconsistent response generation. 
However, in contrast to the rapid development in open-domain dialogue,
few efforts have been made to the task-oriented dialogue direction.
In this paper, we argue that \textit{consistency problem} is more urgent in task-oriented domain.
To facilitate the research, we introduce CI-ToD, a novel dataset for \textbf{C}onsistency \textbf{I}dentification in \textbf{T}ask-\textbf{o}riented \textbf{D}ialog system.
In addition, we not only annotate the single label to enable the model to judge whether the system response is contradictory, but also provide more fine-grained labels (i.e., Dialogue History Inconsistency, User Query Inconsistency and Knowledge Base Inconsistency) to encourage model to know what inconsistent sources lead to it.
Empirical results show that state-of-the-art methods only achieve 51.3\%, which is far behind the human performance of 93.2\%, indicating that there is ample room for improving consistency identification ability.
Finally, we conduct exhaustive experiments and qualitative analysis to comprehend key challenges and provide guidance for future directions.
All datasets and models are publicly available at \url{https://github.com/yizhen20133868/CI-ToD}.
\end{abstract}

\section{Introduction}
\label{Introduction}
\textit{Task-oriented dialogue systems} (ToDs)~\citep{DBLP:journals/pieee/YoungGTW13} aim to achieve user goals such as 
hotel booking and 
restaurant reservation, has gained more attention recently in both academia and industries. 
Over the last few years, two promising research directions in ToDs have emerged.
The first focuses on a pipeline approach, which consists of modularly connected components \cite{wu-etal-2019-transferable,takanobu-etal-2020-multi,peng-etal-2020-shot,li-etal-2020-slot}.
The second direction employs an end-to-end model, which directly takes the sequence-to-sequence (Seq2Seq) model to generate a response from a dialogue history and a corresponding knowledge base (KB) \cite{eric-etal-2017-key,madotto-etal-2018-mem2seq,wen-etal-2018-sequence,qin-etal-2019-entity,wu2019globaltolocal,qin-etal-2020-dynamic}

\begin{figure}[t]
	\centering
	\includegraphics[width=0.44\textwidth]{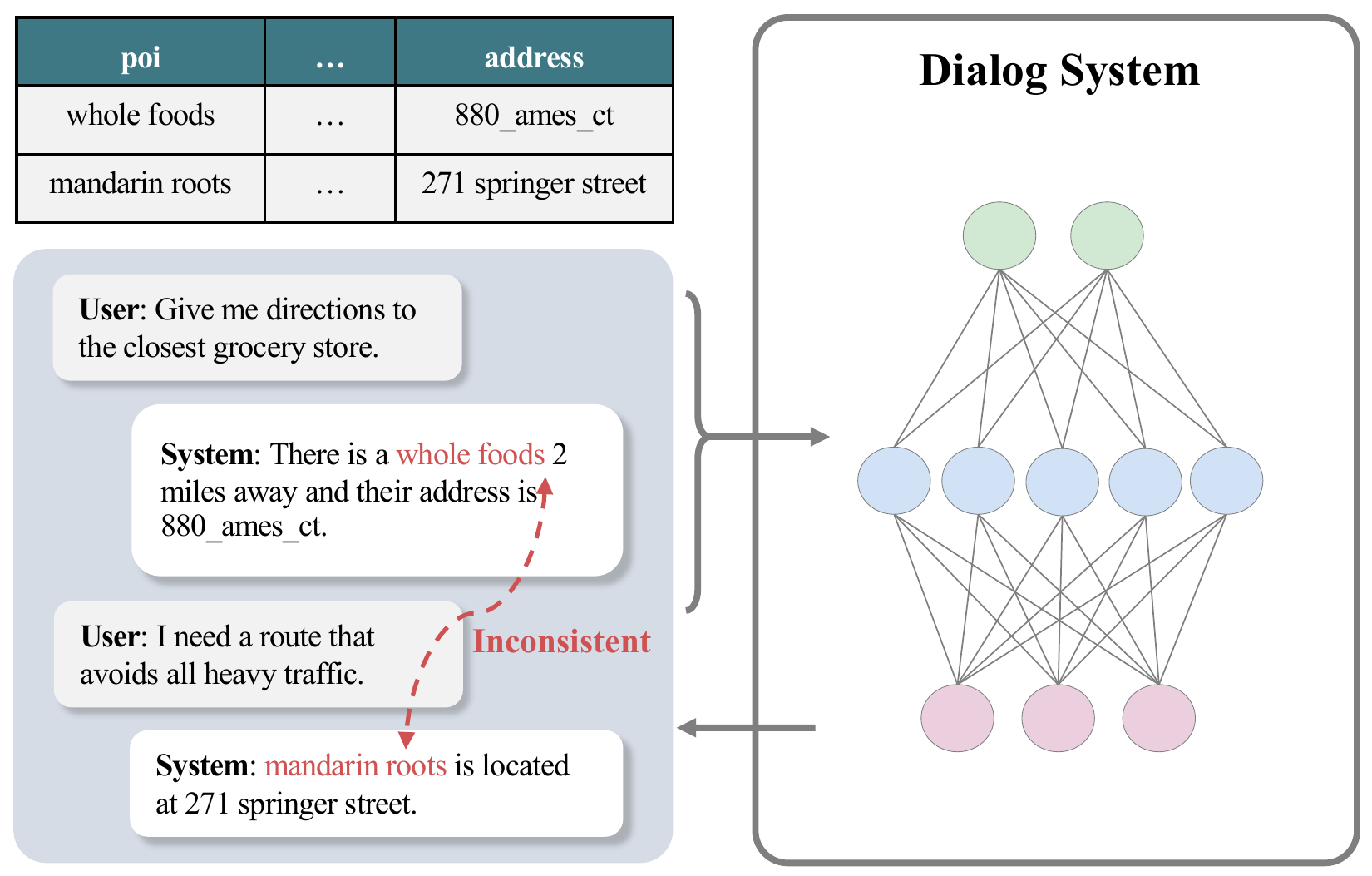}
	\caption{
		A system response generation example by the state-of-the-art end-to-end task-oriented dialogue model \textit{DF-Net} \cite{qin-etal-2020-dynamic}.}
	
	\label{fig:example}
\end{figure}

\newcommand{\tabincell}[2]{\begin{tabular}{@{}#1@{}}#2\end{tabular}}  
\begin{table*}[t] \scriptsize
	\centering
	\begin{adjustbox}{width=1.0\textwidth}
	\begin{tabular}{cccccc}
		\hline
		Dataset & \tabincell{c}{Open Domain/\\Task-Oriented Dialoge System} &\tabincell{c}{External Knowledge} & \tabincell{c}{Multi-turn / Single-turn} & Single Label / Fine-grained Labels \\ 
		\hline
		Dialogue NLI \cite{welleck-etal-2019-dialogue} & Open domain&  \XSolidBold  & Single-Turn & Single Label  \\
			InferConvAI \cite{dziri-etal-2019-evaluating} &Open domain &\XSolidBold  &Multi-Turn & Single Label   \\
		KvPI \cite{song-etal-2020-profile} & Open domain& \CheckmarkBold & Single-Turn  & Single Label  \\
		DECODE \cite{nie-etal-2021-like}& Open domain& \XSolidBold     & Multi-Turn & Single Label   \\
		\hline
		\text{CI-ToD}   & Task-Oriented &\CheckmarkBold   & Multi-Turn      &  \tabincell{c}{Fine-grained Labels (HI, QI and KBI) }   \\ 
		\hline
	
	\end{tabular}
	\end{adjustbox}
	\caption{Comparison between our dataset and other datasets. HI denotes Dialog History Inconsistency; QI denotes User Query Inconsistency; KBI represents Knowledge Base Inconsistency.}
	\label{comparisons}
	\vspace{-10pt}
\end{table*}
In recent years, with the burst of deep neural networks and the evolution of pre-trained language models, 
the research of ToDs has obtained great success.
While the success is indisputable, 
previous work have shown that it's inevitable to generate inconsistent response with the neural-based model, resulting in a contradiction~\cite{welleck-etal-2019-dialogue,song-etal-2020-profile,nie-etal-2021-like}. 
Such contradictions caused by these bots are often jarring, immediately disrupt the conversational flow.
To address the above issue, some work try to improve consistency in dialogue by posing a consistency identification into dialogue.  \citet{welleck-etal-2019-dialogue} made an early step towards performing consistency identification in dialogue agent.
 \citet{nie-etal-2021-like} proposed dialogue contradiction detection task to prevent the system response from being inconsistent with dialogue history.
\citet{song-etal-2020-profile} further proposed a profile consistency identification to consider whether response is consistent with the corresponding profile.
Though achieving the promising performance, the above work were limited to open-domain dialogue.
In this paper, we highlight that \textit{inconsistent generation problems should also be considered in task-oriented dialogue}.
For example, as shown in Figure~\ref{fig:example}, the system expresses about the POI \textit{whole foods} in dialogue history.
However, when we run the state-of-the-art model (\textit{DF-Net}) \cite{qin-etal-2020-dynamic}, the system generate response \textit{``mandarin roots is located at 271 springer street.''}, which incorrectly generates irrelevant POI \textit{mandarin roots}, resulting in contradiction.
This is because neural-based models are a black-box and thus make us hard to explicitly control 
the neural-based dialogue systems
to maintain a consistent response generation.
From the user’s perspective, such inconsistent bots not only fail to meet the requirements of the user but also mislead users to get wrong feedback in the task-oriented domain.
Therefore, it's promising to consider \textit{consistency} problem and detect in advance whether the generated response is consistent in task-oriented dialogue direction.
Unfortunately, there still has been relatively little research on considering consistency identification in task-oriented dialogue due to the the lacking of public benchmarks.

To fill this research gap, we introduce a novel human-annotated dataset CI-ToD: \textbf{C}onsistency \textbf{I}dentification in \textbf{T}ask-\textbf{o}riented \textbf{D}ialog system.
Dialogue data for
CI-ToD is collected from the public dialogue corpora KVRET \cite{eric-etal-2017-key}.
For each final system response in KVRET, we re-write the utterance by crowdsourcing where we deliberately contradict the dialogue history, user query or the corresponding knowledge base (KB).
As shown in Table \ref{comparisons}, compared to the existing consistency identification for dialogue dataset, CI-ToD has the following characteristic:
(1) \textit{Task-oriented Dialogue Domain}. To the best of our knowledge, we are the first to consider dialog consistency in task-oriented dialogue system while the prior work mainly focuses on the open domain dialogue system.
We hope CI-ToD can fill the gap of \textit{consistency identification} in the task-oriented dialogue domain;
(2) \textit{Fine-grained Annotations.} 
We provide not only single annotations of whether each sentence is consistent, but also more fine-grained annotations, which can be used for helping the model analyze what source is causing this inconsistency.

To establish baseline performances on CI-ToD, we evaluate the state-of-the-art pre-trained and non pre-trained models for consistency identification. Experimental results demonstrate a significant gap between machine and human performance, indicating there is ample room for improving consistency identification ability.
In addition, we show that our best consistency identification detector correlates well with human judgements, demonstrating that it can be suitable for use as an automatic metric for checking task-oriented dialogue consistency.
Finally, we perform exhaustive experiments and qualitative analysis to shed light on the challenges that current approaches faced with CI-ToD. 

In summary, our contributions are three-fold: 
\begin{itemize}
	\item We make the first attempt to consider consistency identification in task-oriented dialog and introduce a novel human-annotated dataset CI-ToD to facilitate the research. 
	
	\item We establish various baselines for future work and show well-trained consistency identification model can be served as an automatic metric for checking dialogue consistency.
	\item We conduct exhaustive experiments and qualitative analysis to comprehend key challenges and provide guidance for future CI-ToD work.
\end{itemize}

\section{Problem Formulation}
\label{Background}
\begin{figure*}[t]
	\centering
	\includegraphics[width=1.0\textwidth]{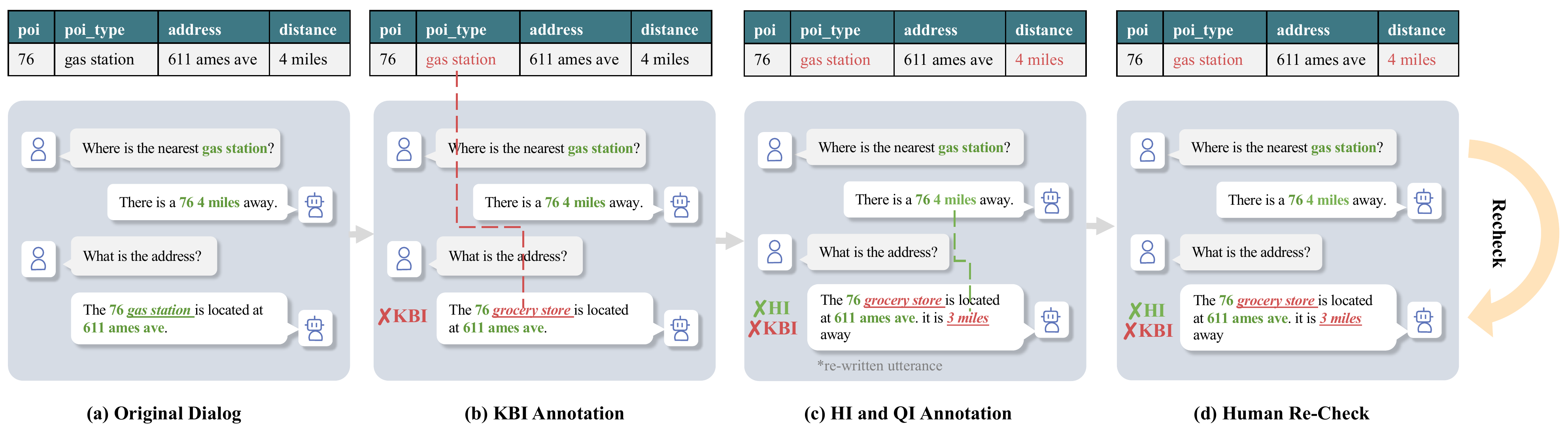}
	\caption{
		The process of CI-ToD construction.}
	\label{fig:construction-step}
\end{figure*}

\begin{figure}[t]
	\centering
	\includegraphics[scale=0.5]{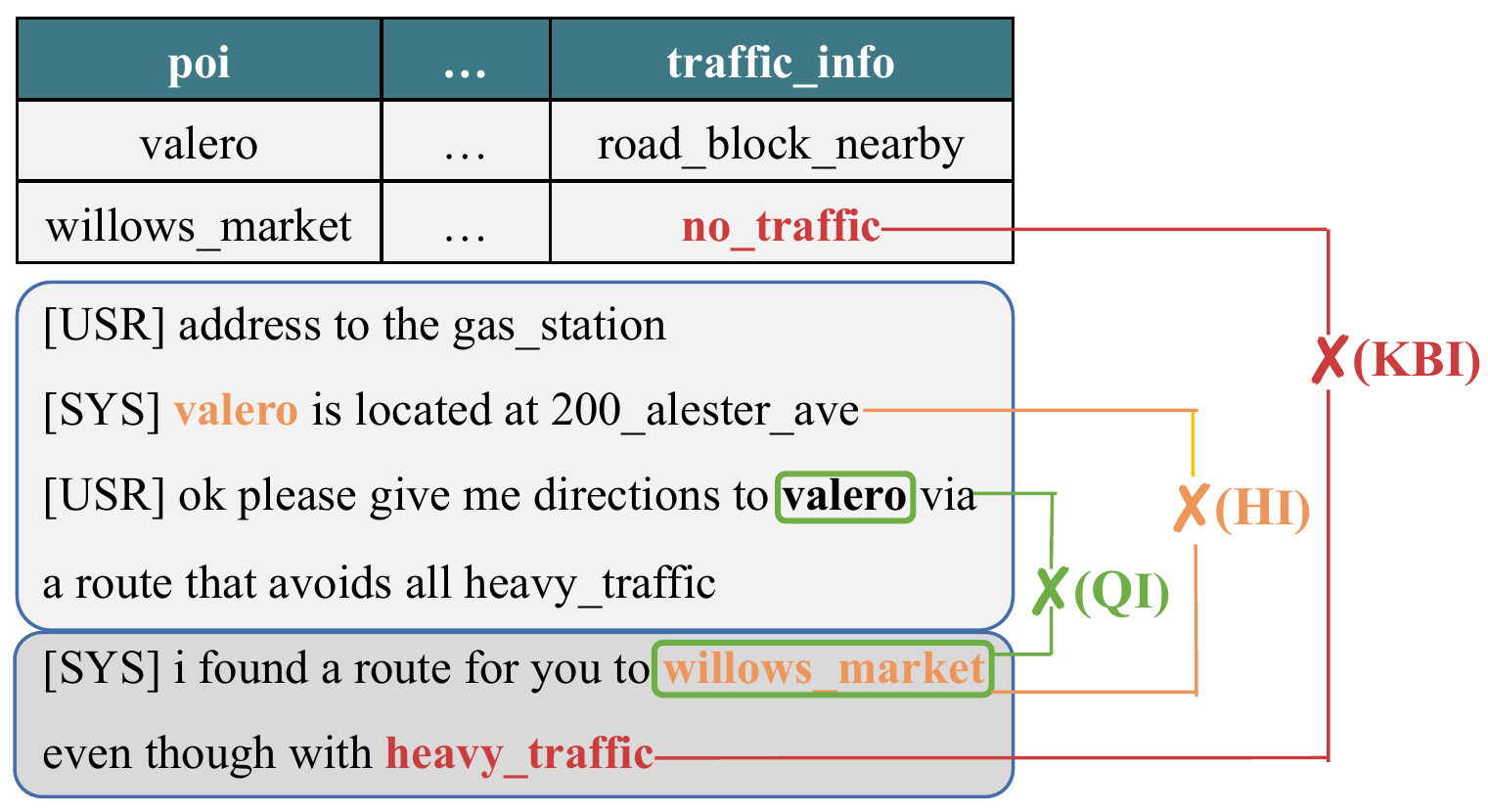}
	\caption{
		Inconsistent types in CI-ToD. Different colors denote different inconsistent types.
	}
	\label{Inconsistency type}
\end{figure}

\begin{figure*}[t]
	\centering
	\includegraphics[width=0.9\linewidth]{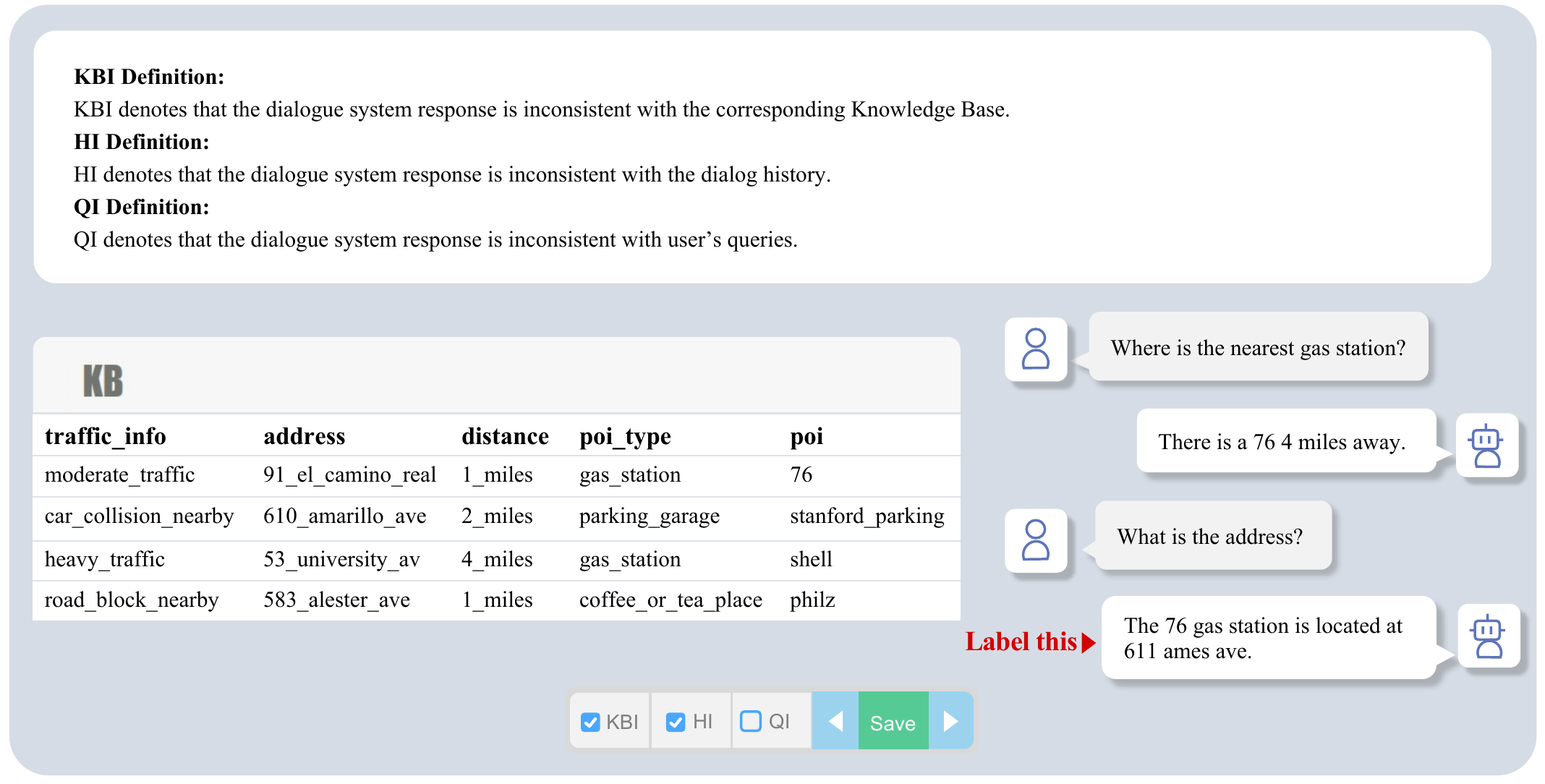}
	\caption{
		The collection interface.
	}
	\label{annotation interface}
\end{figure*}

In our paper, the consistency identification in task-oriented dialogue is formulated as a supervised multi-label classification task, which aims to judge whether the generated system response is inconsistent.
To equip the model with the ability to analyze what the inconsistent sources lead to it, we require the model not only provide the final prediction but also the fine-grained sources including dialogue history, knowledge base (KB) and user's query.
More specifically, given a task-oriented dialogue between a user ($u$) and a system ($s$), the $n$-turned dialogue snippet consists of dialogue history $H$ = $\{(u_{1}, s_{1} ), (u_{2} , s_{2} ), ... , (u_{n-1} , s_{n-1} )\}$, the corresponding knowledge base $KB$, the user query $u_{n}$ and system response $s_{n}$.
More specifically, the task can be defined as:
\begin{eqnarray}
		\mathbf{y} = f ([H, KB, u_{n}], s_{n}), 
\end{eqnarray}
where $f$ denotes the trainable model; $\mathbf{y}$ is an output three-dimension vector, indicating whether the last utterance $s_{n}$ contradicts any previously mentioned dialogue history $H$, user query $u_{n}$ or the corresponding knowledge base $KB$ .

\label{Approach}

\section{Dataset} 
We construct the CI-ToD dataset based on the KVRET dataset and follow four steps: (a) \textit{Data Pre-Processing}, (b) \textit{KBI Construction}, (c) \textit{QI and HI Construction} and (d) \textit{Human Annotation}, which is illustrated in Figure~\ref{fig:construction-step}.
In the following, we first describe the definition of QI, HI and KBI, then illustrate the four construction steps in detail.

\subsection{Inconsistency Types}
As show in Figure~\ref{Inconsistency type}, we give an example to show different inconsistency types, which are illustrated as follows:
\paragraph{User Query Inconsistency (QI)}
QI denotes that the dialogue system response is inconsistent with the current user query. 
Take the dialogue in Figure~\ref{Inconsistency type} for example, in the last turn of dialogue,
user's query is asking about \textit{valero}, while the final system response don't satisfied with user's requirement, showing a route to \textit{willows\_market}, which causes the user query inconsistency.
\paragraph{Dialogue History Inconsistency (HI)}
HI denotes that the dialogue system response is inconsistent with the dialogue history except the current user query. 
Take the dialogue in Figure~\ref{Inconsistency type} for example, 
the previous system response is talking about  \textit{valero} and the user do not change the theme of the dialogue. However,  the final system response turn to discussing about \textit{willows\_market} , causing the dialogue history inconsistency.
\paragraph{Knowledge Base Inconsistency (KBI)}
KBI denotes that the dialogue system response is inconsistent with the corresponding KB, which is an unique challenge in task-oriented dialogue domain. 
Take the dialogue in Figure~\ref{Inconsistency type} for example, the final system response express the traffic\_info of \textit{willows\_market} is \textit{heavy\_traffic}, which is conflict with the corresponding KB (\textit{no\_traffic for willows\_market}).

\subsection{Data Collection and Statistics}
\subsubsection{Step 1 Data Pre-Processing}
We build CI-ToD on existing dialogues KVRET rather than collecting new dialogue from scratch
More specifically, given a $n$-turned dialogue $\{(u_{1}, s_{1} ), (u_{2} , s_{2} ), ... , (u_{n}, s_{n}), KB\}$ for KVRET, we first split it into some sub-dialogues to generate various samples, such as $\{(u_{1}, s_{1} ), KB\}$, $\dots$, $\{(u_{1}, s_{1} ), (u_{2} , s_{2} ), ... , (u_{n-1}, s_{n-1}), KB\}$ and $\{(u_{1}, s_{1} ), (u_{2} , s_{2} ), ... , (u_{n}, s_{n}), KB\}$.
In addition, to ensure the system response is informative, we filter these general response, such as \textit{``Thanks''} and \textit{``You are welcome''}.
Finally, we obtain the pre-processed dialogues.

\begin{table}[t] 
	\centering
	\begin{adjustbox}{width=0.35\textwidth}
		\begin{tabular}{c|c}
			& CI-ToD \\
			\hline
			\# Domain & 3 \\
			\# Training Dialogues & 2,553\\
			\# Validation Dialogues & 319\\
			\# Test Dialogues & 318\\
			\# Avg. Utterances per Dialogue & 3.693\\
			\# Inconsistency ratio& 0.648\\
			\# KBI ratio & 0.521 \\
			\# QI ratio & 0.485\\
			\# HI ratio & 0.214\\
			\hline
		\end{tabular}
	\end{adjustbox}
	\caption{Data statistics of CI-ToD.}
	\label{table:statistic}
	\vspace{-10pt}
\end{table}

\subsubsection{Step 2 KBI Annotation}
Given the pre-processed dialogues, we first construct KBI for each dialogue.
KBI denotes that the final system response is inconsistent with the corresponding KB.
We simply replace the knowledge entity value to construct KBI automatically.

More specifically, for each knowledge value in the system response, 
 we sample specific entities from the whole KB to replace the selected slot and ensure that the sampled KB entity is different with the selected value. By this means, the constructed response is inconsistent with the corresponding KB.
 For example, as shown in Figure~\ref{fig:construction-step}(b), we replace the entity \textit{``gas station''} with \textit{``grocery store''}, which resulting in KBI (the corresponding KB is (poi\_type for \textit{gas station})).

\subsubsection{Step 3 QI and HI Annotation}
In this section, we show how we generate QI and HI.
Since this require us to have a deep understanding for the corresponding user's query and dialogue history, constructing a system response with QI or HI is non-trivial, 
To address this issue, we achieve this by human efforts. 
We hire a human annotation team\footnote{All annotators are undergraduates from university in China, who are familiar with English language. (pass the College English Test (CET-6), one of the hardest English level exams in China.) } to (1) randomly assign a sample with QI or HI and re-write each response to make it inconsistent with user query or dialogue history, and (2) check whether each written response is fluent or not by three extra annotators.

\begin{figure*}[t]
	\centering
	\includegraphics[scale=0.4]{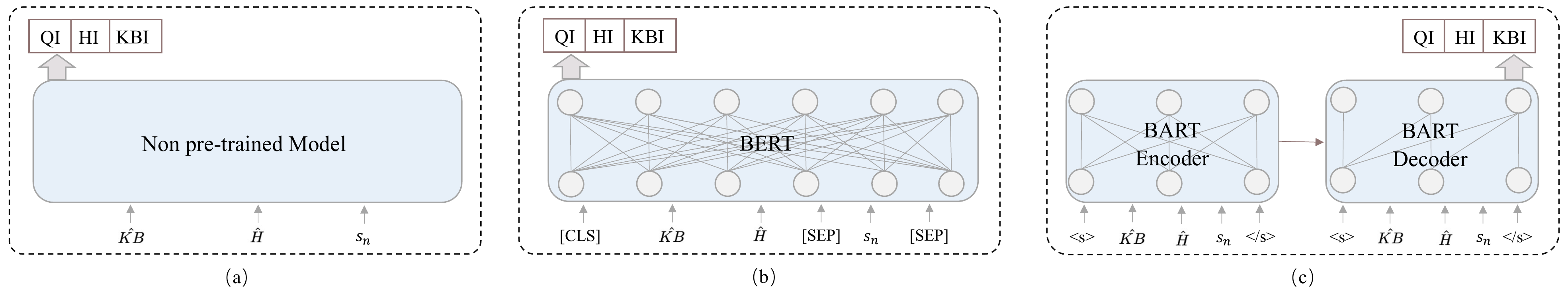}
	\caption{
		The model structure of non pre-trained model (a) and pre-trained model (b and c). 
	}
	\label{fig:pre-trained-model}
\end{figure*}
\subsubsection{Step 4 Human Re-Check}
In the final step, we will re-check the fine-grained inconsistent information with human efforts, including QI, HI and KBI. To ensure quality, each sample is annotated by three people, and the annotation process lasts nearly three months.
Figure~\ref{annotation interface} shows the annotation user interface.

The detailed statistics of CI-ToD is summarized in Table~\ref{table:statistic}. 
The percentage of inconsistency has exceeded 50\%, indicating that CI-ToD is challenging.

\subsubsection{Quality Control}
To control the quality of the annotated dataset, we introduce different verification methods: \\
(1) \textbf{Onboarding Test:} Each annotator will have an advance annotation test, where each annotator will first annotate 100 samples and 3 experts check their annotation results.
Finally, only those who achieves 80\% annotation performance can conduct the following annotation work;
(2) \textbf{Double Check} We randomly sampled 1,000 samples from the final annotated dataset and ask two new annotators to annotate the inconsistent information. Following \cite{bowman-etal-2015-large}, we calculated the Fleiss’ Kappa among the previous labels and two new labels and obtained a kappa of 0.812, which means almost perfect agreement \cite{Landis77}. 


\section{Models}
In this section, we establish several strong baseline methods using the state-of-the-art non pre-trained models ($\S\ref{sec:model-non-pre-trained}$) and pre-trained models ($\S\ref{sec:model-pre-trained}$).
Since multi-task framework has obtained remarkable success on various NLP tasks \cite{fan2021multi, qin-etal-2019-stack, Qin_Che_Li_Ni_Liu_2020, liang2020aspect, xu-etal-2021-xlpt,qin2021cointeractive}, we adopt a vanilla multi-task framework to simultaneously perform QI, HI, and KBI, which has the advantage of extracting the shared knowledge across three tasks.

For both pre-trained models and non-pre-trained models, we introduce delimiter tokens  \texttt{[SOK]}, \texttt{[USR]} and \texttt{[SYS]} to signal the beginning of KB, user utterance and system response, respectively, aiming to learn to distinguish the role of KB, user and system behavior in multi-turn dialogues. 
Specifically, the input of KB is denoted as $\hat {KB}$ = "$[SOK]\ $KB$\ [EOK]$" while input of H is defined as $\hat {H}$ = "$[USR]\ u_1\ [SYS]\ s_1\ ...\ [USR]\ u_n$".

\subsection{Non Pre-trained Models}\label{sec:model-non-pre-trained}
In this approach, we simply concatenate all the previous utterances in the dialogue history and the corresponding KB to form a single textual context, which is shown in Figure~\ref{fig:pre-trained-model}. 
For KB representation, we format each knowledge entity into "column name, cell value" pairs instead of "subject, relation, object" triples to save length space. KB representation for ToDs is actually an important issue which is mentioned in our challenge section.
Then, we apply $f_\text{non}$ as the non pre-trained models to obtain the final prediction, which is defined as:
\begin{eqnarray}
\mathbf{y} = f_\text{non} ([\hat{KB}, \hat{H}, u_{n}], s_{n}) .
\end{eqnarray}

In our work, we explore some state-of-the-art non pre-trained models including: \textit{ESIM} \cite{chen-etal-2017-enhanced}, \textit{InferSent} \cite{conneau-etal-2017-supervised} and \textit{RE2} \cite{yang-etal-2019-simple}.

\subsection{Pre-trained Models}\label{sec:model-pre-trained}
We investigate several state-of-the-art  BERT-based and BART models, which are illustrated in Figure \ref{fig:pre-trained-model}.
Given a dialogue $\{{(u_{1}, s_{1})}, \dots, {(u_{n}, s_{n})}, KB\}$, for BERT-based models, following \cite{chen2020tabfact}, the input can be denoted as  (\texttt{[CLS]}, $\hat {KB}$, $\hat H$, \texttt{[SEP]}, $s_{n}$, \texttt{[SEP]}), where \texttt{[CLS]} and \texttt{[SEP]} are special symbol for classification token and separator token.
After pre-training model encoding, the last layer's hidden representation $\mathbf{h}_\texttt{CLS}$ from the \texttt{[CLS]} token is used for classification, which can be defined as:
\begin{eqnarray}
\mathbf{y}= \text{Softmax} (\mathbf{W}\mathbf{h}_\texttt{CLS} + \mathbf{b}),
\end{eqnarray}
where  $\mathbf{W}$ and $\mathbf{b}$ are the trainable parameters. 

For BART, we feed the same sequence to both the encoder and the decoder, using the last hidden state for classification.
The class that corresponds to the highest probability is chosen as model prediction, which is illustrated in Figure~\ref{fig:pre-trained-model}(b).

More specifically, we explore \texttt{BERT}
\cite{devlin-etal-2019-bert}, \texttt{RoBERTa} \cite{liu2019roberta}, \texttt{XLNet} \cite{yang2020xlnet}, \texttt{Longformer} \cite{beltagy2020longformer} and \texttt{BART} \cite{lewis-etal-2020-bart}.

\subsection{Training Objective}
The training objective is the binary cross-entropy loss, which is defined as:
\begin{equation}
\mathcal{L} \triangleq -\sum_{i=1}^{3} \left(\hat{y}_{i} \log \left({y}_{i}\right)+\left(1-\hat{y}_{i}\right) \log \left(1-{y}_{i}\right)\right) \,
\end{equation}
where $y_{i}$ is the predicted score between 0 and 1 while $\hat {y}_{i}$ is the gold label for the $i$ inconsistent type.

\begin{table*}[t] \small
    \centering
    	\begin{adjustbox}{width=1.0\textwidth}
    \begin{tabular}{c|c|ccc|c} 
    	
    \hline
    Baseline category  & Baseline method & QI F1 & HI F1 & KBI F1  & Overall Acc \\
    \hline
		
     \multirow{3}{*}{\shortstack[c]{Non pre-trained Model}}    
		&  ESIM \cite{chen-etal-2017-enhanced} & 0.512 & 0.164 & 0.543  
& 0.432 \\
		&  InferSent \cite{conneau-etal-2017-supervised}  & 0.557 & 0.031 & 0.336 
& 0.356\\
     &  RE2 \cite{yang-etal-2019-simple}  & 0.655 & 0.244 & 0.739  
& 0.481\\
      \hline
    \multirow{5}{*}{\shortstack[c]{Pre-trained Model}}
		&  BERT \cite{devlin-etal-2019-bert} &0.691&	{\bf 0.555} &	0.740  
& 0.500 \\
     &  RoBERTa \cite{liu2019roberta} &0.715&	0.472&	0.715
&0.500 \\ 
		&  XLNet \cite{yang2020xlnet}  &0.725&	0.487&	0.736 
&	0.509 \\
		&  Longformer \cite{beltagy2020longformer} &0.717&	0.500 &	0.710 
&0.497 \\
		&  BART \cite{lewis-etal-2020-bart} &{\bf 0.744}&	0.510&{\bf 0.761}
&{\bf 0.513} \\    
    \hline
	\multirow{1}{*}{\shortstack[c]{Human}}
    	&  Human Performance& 0.962 & 0.805 & 0.920 
& 0.932 \\
		\hline
    \end{tabular}
\end{adjustbox}
    \caption{Comparison of varying approaches on CI-ToD dataset.
    }
    \label{tab:main_performance}
    \vspace{-5pt}
\end{table*}

\section{Experiments}
\label{experiments}
\subsection{Implementation Details}

For pre-trained models the batch size we use in our framework is selected from $\{4, 8\} $ and learning rate is selected from  $\{5e^{-6}\}$ to $\{2e^{-5}\}$ with a step of $\{1e^{-6}\}$.
We set the max length to 512 tokens for all models except \texttt{Longformer}, of which 3,000 tokens are the max length we take. 
For the non pre-trained models, we adopt the suggested hyper-parameters in their open-sourced code.
All experiments are conducted at TITAN Xp and Tesla V100 GPUs. For all experiments, we select model which performs best on the development set and evaluate it on the test set.

\subsection{Evaluation}
We adopt overall accuracy (Overall Acc) to evaluate model's performance, measuring the ratio of sample for which both QI, HI and KBI are predicted correctly.
Furthermore, to give more detailed analysis, we also calculate F1 score on the \textsc{QI}, \textsc{HI} and \textsc{KBI} labels.

\subsection{Human Performance}
To measure human performance on the CI-ToD dataset, we 
ask three experts to judge each sample from dataset.
Only if the results of the three experts are consistent, we consider this sample is predicted correctly by human.
The human performance is shown in the last row of Table~\ref{tab:main_performance}.

\subsection{Main Results}
 Table~\ref{tab:main_performance} shows the results of the models discussed in the previous section. 
 
From the results, we have the two interesting observations: (1) The human performance is 93.2\%. In contrast, all of the non pre-trained and pre-trained models perform significantly worse than humans, demonstrating that there is ample room for improving consistency ability in the task-oriented dialogue; (2) Pretrained models outperform all non-pre-trained models in CI-ToD, which is consistent with results in other literature \cite{talmor-etal-2019-commonsenseqa}. We think that knowledge learned from pre-training can benefitial to consistency identification.

\subsection{Qualitative Analysis}

\subsubsection{Performance Across Different Consistency Types}
We compare human performance and model performance across different consistency types.
The results are shown in Table~\ref{tab:main_performance}. We can observe that humans are good at deciding the all consistency types, indicating that it's easy for human to detect whether a dialogue is consistent because human have a strong reasoning ability.
In contrast, we find that the best pre-trained model (\texttt{BART}) obtains the worst results on \textrm{HI} type compared with other types (\textit{QI and KBI}). This is because that correctly detecting \textrm{HI} rely on the dialogue context information which faces the challenges of coreference resolution. We will discuss it in details in Section~\ref{sec:challenges}.

\subsubsection{Context Ablation Study}
\begin{figure}[t]
	\centering
	\includegraphics[width=0.45\textwidth]{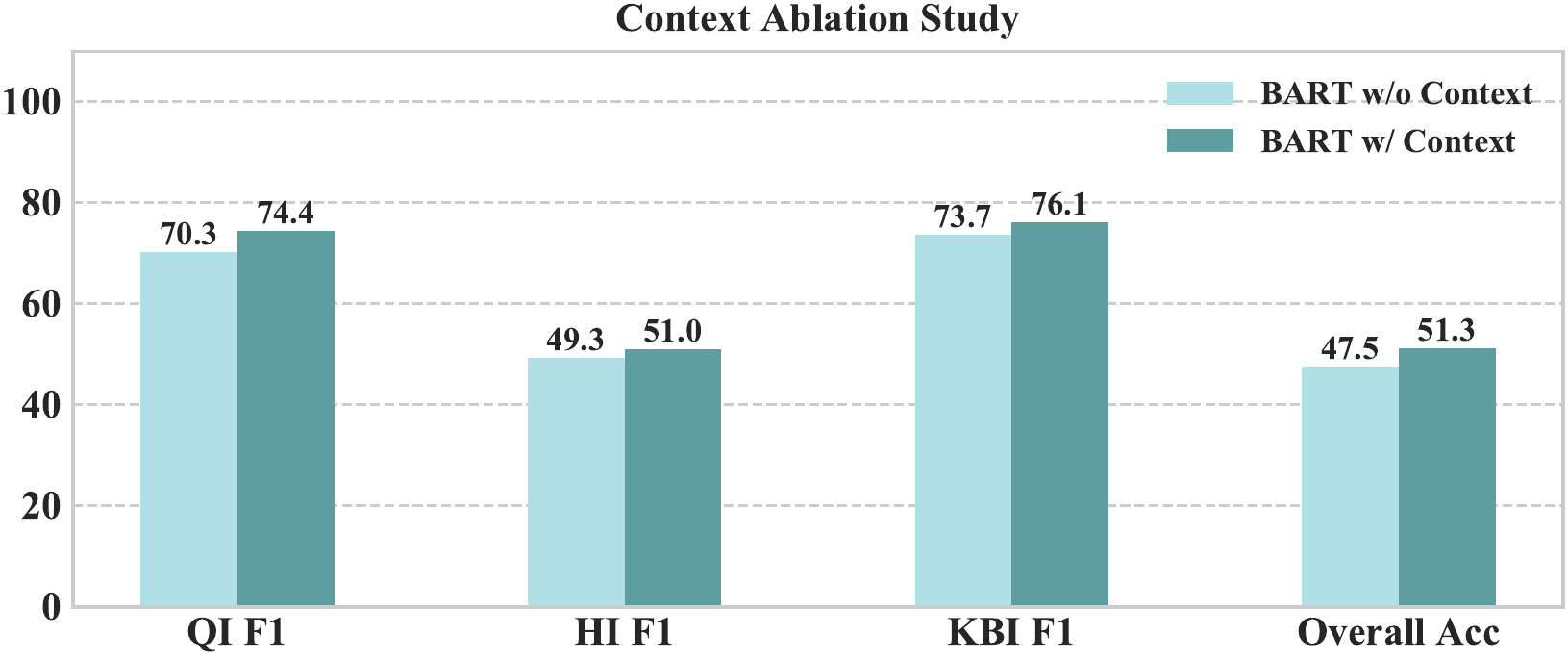}
	\caption{
		Context Ablation Study
	}
	\label{fig:context_ablation_study}
\end{figure}

In this section, we analyze the impact of context on final performance. 
More specifically, we conduct  experiments by removing the corresponding dialogue contextual information and only keeping the final user query. Figure~\ref{fig:context_ablation_study} shows the results of \texttt{BART} without contextual information. We observe that our model drops in all consistency types.
This is because dialogue context can help model to understand the whole dialogue topic, which is useful to the consistency detection.

\subsubsection{Multi-Task Training vs. Separate Task Training}

\begin{figure}[t]
	\centering
	\includegraphics[width=0.45\textwidth]{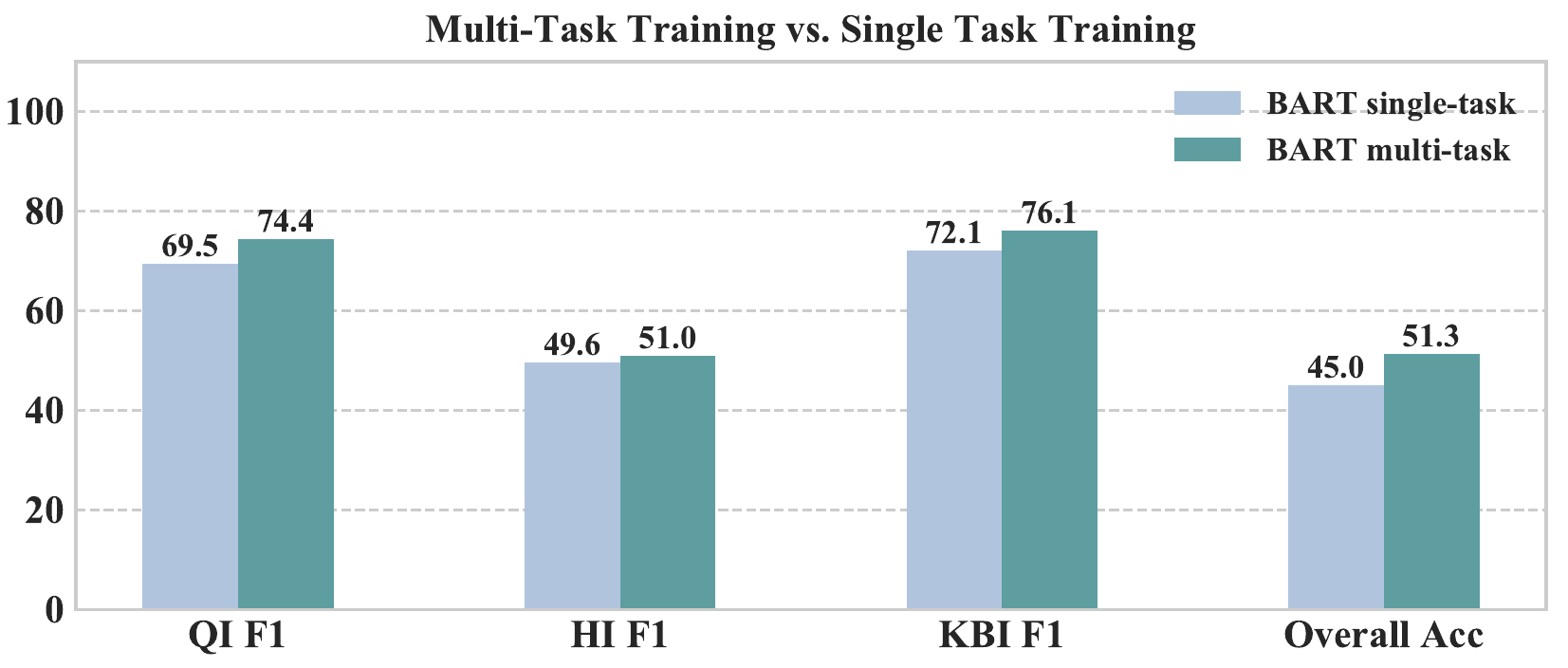}
	\caption{
		Multi-Task Training vs. Single Task Training
	}
	\label{fig:parameter_sharing_ablation_study}
\end{figure}

In this section, we explore the effectiveness of the proposed multi-task framework.
In particular, we conduct separate training setting where we use the \texttt{BART} to perform each task prediction (\textrm{QI}, \textrm{HI} and \textrm{KBI}) separately.
The comparison results are shown in Figure~\ref{fig:parameter_sharing_ablation_study}, we can observe that model with multi-task training outperforms separate task training paradigm in all metrics, which indicates that \textrm{QI}, \textrm{HI} and \textrm{KBI} tasks are correlated, and thus modeling the correlation across tasks can improve performance.

\subsubsection{Using CI-ToD as an Automatic Metric}

\begin{figure}[t]
	\centering
	\includegraphics[width=0.45\textwidth]{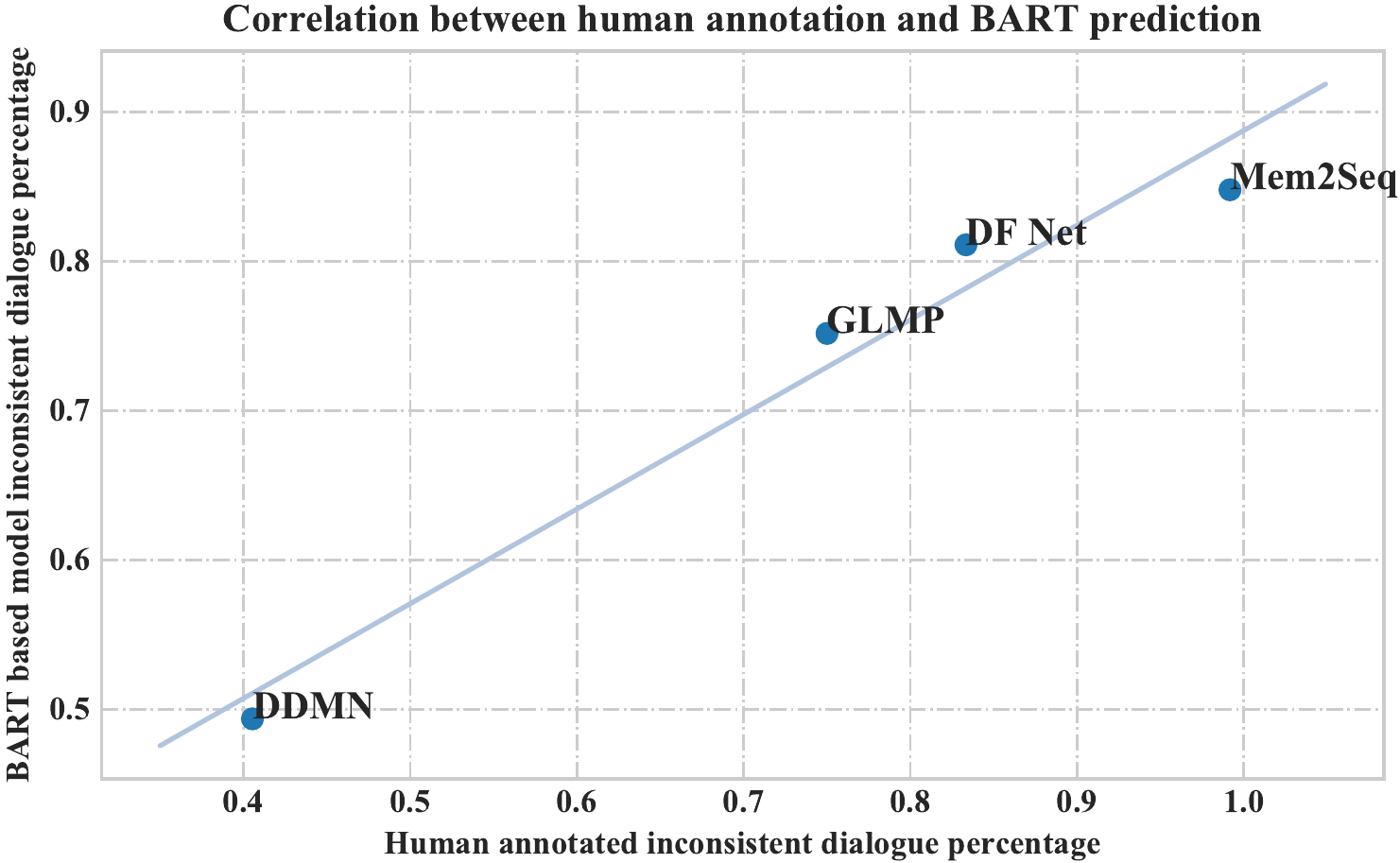}
	\caption{
		Correlation between human annotation and BART prediction
	}
	\label{inconsistency_correlation}
\end{figure}

In this section, we want to further investigate whether it can judge the quality of the utterances by different task-oriented dialogues and be used as an automatic metric checking generation consistency.
We compare the overall accuracy of the well-trained best model \texttt{BART} with the contradiction rate by human judgements on the utterances generated by different models.
In particular, we explore the state-of-the-art end-to-end task-oriented dialogue models (\texttt{Mem2seq} ~\cite{madotto-etal-2018-mem2seq}, \texttt{GLMP}~\cite{wu2019globaltolocal}, \texttt{DF-Net}~\cite{qin-etal-2020-dynamic}, \texttt{DDMN}~\cite{wang-etal-2020-dual}). 
The results are shown in Figure~\ref{inconsistency_correlation} and we can see that the scores are positively correlated with human judgments, with a Pearson correlation coefficient of 0.9.
This demonstrates the proposed consistency identification model can be used as a automatic metric to evaluate consistency in task-oriented dialogues.
\begin{figure*}[ht]
	\centering
	\includegraphics[width=0.95\textwidth]{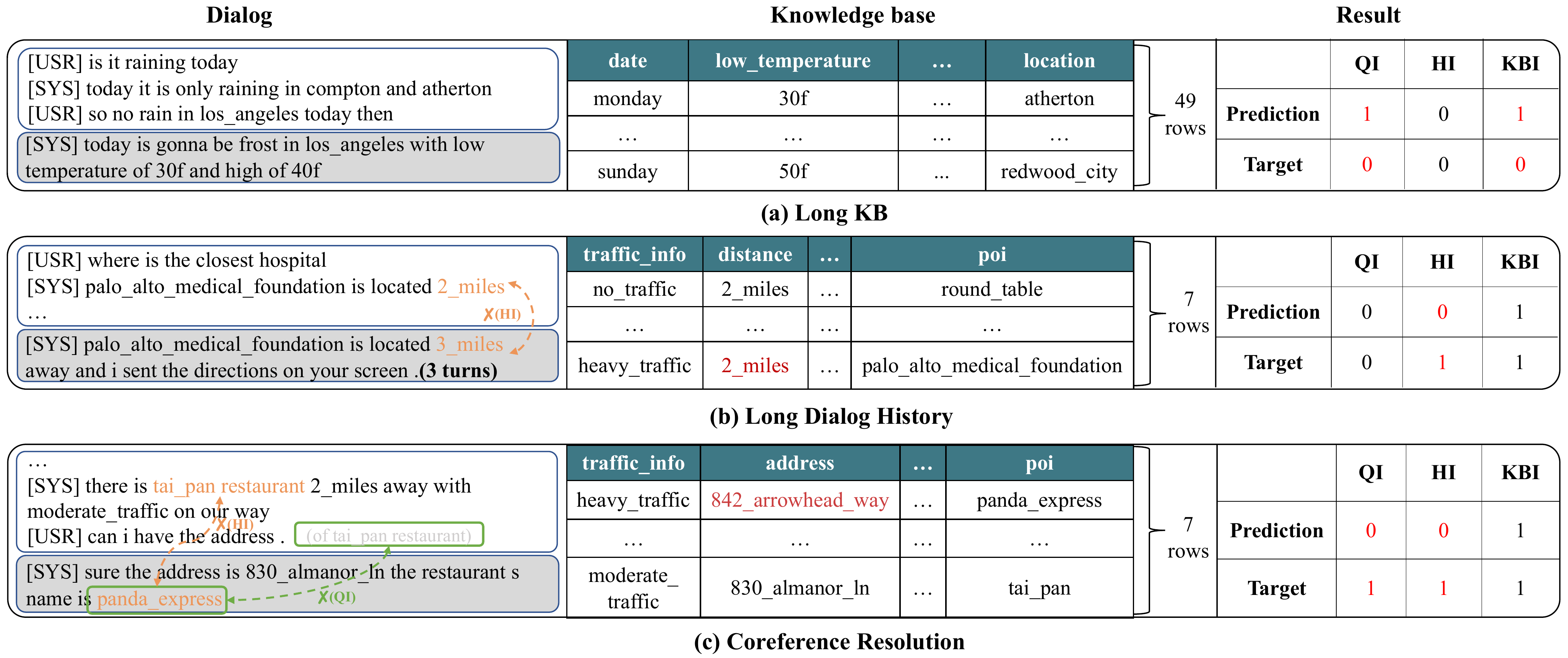}
	\caption{
		Error type in CI-ToD baseline model's prediction.
	}
	\label{fig:error-type}
\end{figure*}

\subsection{Error Analysis}
In this section, we empirically divide all the error samples generated with \texttt{BART} into three categories, which are shown in Figure~\ref{fig:error-type}.
\paragraph{Long KB.} When the KB is relatively large, it contains a lot of redundant information which is irrelevant to the current conversation. 
This redundant information will become noise in the process of model learning and simply flatting the KB into a sequence can not effectively modeling the relevant information.
 For example, as shown in Figure \ref{fig:error-type}, when the KB is large (49 rows), \texttt{BART} predicts the \textrm{KBI} as 1 incorrectly.
 
\paragraph{Long Dialog History.} When the dialogue history is too long, it may contain some noise information. As shown in Figure \ref{fig:error-type}, there are three rounds of dialogues at this time, the system expresses \textit{``palo\_alto\_medical\_foundation is located 2\_miles away''} at first round of dialogue while describes \textit{``palo\_alto\_medical\_foundation is located 3\_miles away''} at last turn, causing the HI due to the irrelevant middle context.

\paragraph{Coreference Resolution.} When there are some implicit or explicit references in the dialogue, it is necessary to resolve the references to restore the intention of the conversation, which greatly increases the difficulty of the model to predict the types of inconsistencies. For example, in Figure \ref{fig:error-type}, the last round of the user's query \textit{``can i have the address''} does not clearly indicate a specific object, which confuses model to predict the \textrm{QI} and \textrm{HI} as 0 incorrectly.
Actually, by resolving the implicit reference according to the dialogue history, we can know that the reference object of the user's current problem is \textit{``tai\_pan restaurant''}, which helps model to obtain correct results.

\subsection{Challenges}\label{sec:challenges}
Based on above analysis, we summarize the current challenges faced by the consistency detection task:
\paragraph{KB Representation.} The corresponding Knowledge base is the relational database, which has high-order structure information presented in the original knowledge graph.
How to modeling the structure information in the relational knowledge base rather than simply flattening the KB is an interesting research question to investigate.
In addition, since the size of KB is relatively big,
how to effectively modeling relevant KB information rather than injecting noisy is another challenge to explore.

\paragraph{Effectively Context Modeling.}
Since some dialogue has extreme long histories, not all context information have a positive influence for the final performance. 
How to effectively model the long-distance dialogue history and filter irrelevant information is an interesting research topic.
\paragraph{Coreference Resolution.} There are multiple coreference resolution in a dialogue, which
will result in ambiguity in the user's query, making it difficult for model to predict the consistency label correctly.
Thus, how to explicitly conduct coreference resolution to help the consistency detection is an important research question.
\paragraph{Explicit Joint Learning.}
Though achieving promising performance based on the multi-task training paradigm, the prior work did not “explicitly” model the relationships between the different tasks ({HI}, {QI} and {KBI} task); instead, it adopted shared parameters to  “implicitly” model the correlation.
However, simply relying on a set of shared parameters cannot make a full interaction to achieve desirable results \cite{qin-etal-2019-stack, Qin_Che_Li_Ni_Liu_2020}.
Thus, how to explicitly modeling the correlation between {HI}, {QI} and {KBI} to directly control information flow still deserves to be explored.

\section{Related Work}
\label{related work}
This work is related to the considering consistency in open-domain dialogue.
In recent years, several personalized dialogue datasets have been introduced, such as PersonaChat \cite{zhang-etal-2018-personalizing} and PersonalDialog \cite{zheng2020personalized}.
These datasets are able to implicitly consider the consistency in dialogue generation, but fail to explicitly teach the model to judge whether the generated system response is consistent.

Another series of related work explicity improve consistency in dialogue.
To this end, some benchmarks have been proposed to promote the research. 
\citet{welleck-etal-2019-dialogue} made an early step towards reducing the dialogue consistency identification to natural language inference (NLI).
\citet{dziri-etal-2019-evaluating} presented a novel paradigm for evaluating the coherence of dialogue systems by using state-of-the-art entailment techniques and build a synthesized dataset InferConvAI geared toward evaluating consistency in dialogue systems. 
\citet{nie-etal-2021-like} introduced the DialoguE COntradiction DEtection task (DECODE) and a new conversational dataset containing contradictory dialogues, aiming to evaluate the ability to detect contradictory.
\citet{song-etal-2020-profile} proposed a KvPI dataset and profile consistency identification task for open-domain dialogue agents to further evaluate whether the system response is inconsistent with the corresponding profile information.
Compared with their work that mainly focus on the open-domain dialogue direction, we aim to fill the gap of consistency identification in task-oriented dialogue systems. Furthermore, we introduce a human-annotated dataset to this end. Besides, we provide some key challenges and future directions to facilitate further research.

\section{Conclusion}
\label{conclusion}
We studied consistency identification in task-oriented dialogue and introduced a new human-annotated dataset CI-ToD.
Further, we analyzed the problems of CI-ToD through extensive experiments and highlight the key challenges of the task. 
We hope CI-ToD can facilitate future research on consistency identification in task-oriented dialogue.

\section*{Acknowledgements}
This work was supported by the National Key R\&D
Program of China via grant 2020AAA0106501 and
the National Natural Science Foundation of China
(NSFC) via grant 61976072 and 61772153. This
work was also supported by the Zhejiang Lab’s
International Talent Fund for Young Professionals.

\section*{Ethical Considerations}
\paragraph{Data Access.}
We collected our data from KVRET dataset \cite{eric-etal-2017-key}.
This dataset is an open-source dataset free for academic research. 

\paragraph{Annotation Platform Construction.}
The annotation interface is built by authors based on open-resource JAVA framework and HTML. The server for collecting and storing annotation was rent from Alibaba Cloud using our funding.

\paragraph{Dataset Collection Process.}
For the annotation, we ﬁrst launch interviews of the task introduction with 100 example questions, which is paid as \$20, for them to try a few examples to get informed and familiar with the task on onboarding test process. Then during annotation process, annotators were paid \$15.0 per hour, and the total human-hours we cost are about 300 hours. After annotation, the authors re-check those examples with mismatched tags, which cost about another 20 hours


\bibliography{anthology,custom}

\begin{thebibliography}{36}
\expandafter\ifx\csname natexlab\endcsname\relax\def\natexlab#1{#1}\fi

\bibitem[{Beltagy et~al.(2020)Beltagy, Peters, and
  Cohan}]{beltagy2020longformer}
Iz~Beltagy, Matthew~E. Peters, and Arman Cohan. 2020.
\newblock \href {http://arxiv.org/abs/2004.05150} {Longformer: The
  long-document transformer}.

\bibitem[{Bowman et~al.(2015)Bowman, Angeli, Potts, and
  Manning}]{bowman-etal-2015-large}
Samuel~R. Bowman, Gabor Angeli, Christopher Potts, and Christopher~D. Manning.
  2015.
\newblock \href {https://doi.org/10.18653/v1/D15-1075} {A large annotated
  corpus for learning natural language inference}.
\newblock In \emph{Proceedings of the 2015 Conference on Empirical Methods in
  Natural Language Processing}, pages 632--642, Lisbon, Portugal. Association
  for Computational Linguistics.

\bibitem[{Chen et~al.(2017)Chen, Zhu, Ling, Wei, Jiang, and
  Inkpen}]{chen-etal-2017-enhanced}
Qian Chen, Xiaodan Zhu, Zhen-Hua Ling, Si~Wei, Hui Jiang, and Diana Inkpen.
  2017.
\newblock \href {https://doi.org/10.18653/v1/P17-1152} {Enhanced {LSTM} for
  natural language inference}.
\newblock In \emph{Proceedings of the 55th Annual Meeting of the Association
  for Computational Linguistics (Volume 1: Long Papers)}, pages 1657--1668,
  Vancouver, Canada. Association for Computational Linguistics.

\bibitem[{Chen et~al.(2020)Chen, Wang, Chen, Zhang, Wang, Li, Zhou, and
  Wang}]{chen2020tabfact}
Wenhu Chen, Hongmin Wang, Jianshu Chen, Yunkai Zhang, Hong Wang, Shiyang Li,
  Xiyou Zhou, and William~Yang Wang. 2020.
\newblock \href {http://arxiv.org/abs/1909.02164} {Tabfact: A large-scale
  dataset for table-based fact verification}.

\bibitem[{Conneau et~al.(2017)Conneau, Kiela, Schwenk, Barrault, and
  Bordes}]{conneau-etal-2017-supervised}
Alexis Conneau, Douwe Kiela, Holger Schwenk, Lo{\"\i}c Barrault, and Antoine
  Bordes. 2017.
\newblock \href {https://doi.org/10.18653/v1/D17-1070} {Supervised learning of
  universal sentence representations from natural language inference data}.
\newblock In \emph{Proceedings of the 2017 Conference on Empirical Methods in
  Natural Language Processing}, pages 670--680, Copenhagen, Denmark.
  Association for Computational Linguistics.

\bibitem[{Devlin et~al.(2019)Devlin, Chang, Lee, and
  Toutanova}]{devlin-etal-2019-bert}
Jacob Devlin, Ming-Wei Chang, Kenton Lee, and Kristina Toutanova. 2019.
\newblock \href {https://doi.org/10.18653/v1/N19-1423} {{BERT}: Pre-training of
  deep bidirectional transformers for language understanding}.
\newblock In \emph{Proceedings of the 2019 Conference of the North {A}merican
  Chapter of the Association for Computational Linguistics: Human Language
  Technologies, Volume 1 (Long and Short Papers)}, pages 4171--4186,
  Minneapolis, Minnesota. Association for Computational Linguistics.

\bibitem[{Dziri et~al.(2019)Dziri, Kamalloo, Mathewson, and
  Zaiane}]{dziri-etal-2019-evaluating}
Nouha Dziri, Ehsan Kamalloo, Kory Mathewson, and Osmar Zaiane. 2019.
\newblock \href {https://www.aclweb.org/anthology/W19-3646} {Evaluating
  coherence in dialogue systems using entailment}.
\newblock In \emph{Proceedings of the 2019 Workshop on Widening NLP}, pages
  146--148, Florence, Italy. Association for Computational Linguistics.

\bibitem[{Eric et~al.(2017)Eric, Krishnan, Charette, and
  Manning}]{eric-etal-2017-key}
Mihail Eric, Lakshmi Krishnan, Francois Charette, and Christopher~D. Manning.
  2017.
\newblock \href {https://doi.org/10.18653/v1/W17-5506} {Key-value retrieval
  networks for task-oriented dialogue}.
\newblock In \emph{Proceedings of the 18th Annual {SIG}dial Meeting on
  Discourse and Dialogue}, pages 37--49, Saarbr{\"u}cken, Germany. Association
  for Computational Linguistics.

\bibitem[{Fan et~al.(2021)Fan, Yuan, Gui, Zhang, and Xu}]{fan2021multi}
Chuang Fan, Chaofa Yuan, Lin Gui, Yue Zhang, and Ruifeng Xu. 2021.
\newblock Multi-task sequence tagging for emotion-cause pair extraction via tag
  distribution refinement.
\newblock \emph{IEEE/ACM Transactions on Audio, Speech, and Language
  Processing}.

\bibitem[{Landis and Koch(1977)}]{Landis77}
J.~Richard Landis and Gary~G. Koch. 1977.
\newblock The measurement of observer agreement for categorical data.
\newblock \emph{Biometrics}, 33(1).

\bibitem[{Lewis et~al.(2020)Lewis, Liu, Goyal, Ghazvininejad, Mohamed, Levy,
  Stoyanov, and Zettlemoyer}]{lewis-etal-2020-bart}
Mike Lewis, Yinhan Liu, Naman Goyal, Marjan Ghazvininejad, Abdelrahman Mohamed,
  Omer Levy, Veselin Stoyanov, and Luke Zettlemoyer. 2020.
\newblock \href {https://doi.org/10.18653/v1/2020.acl-main.703} {{BART}:
  Denoising sequence-to-sequence pre-training for natural language generation,
  translation, and comprehension}.
\newblock In \emph{Proceedings of the 58th Annual Meeting of the Association
  for Computational Linguistics}, pages 7871--7880, Online. Association for
  Computational Linguistics.

\bibitem[{Li et~al.(2020)Li, Yao, Qin, Che, Li, and Liu}]{li-etal-2020-slot}
Yangming Li, Kaisheng Yao, Libo Qin, Wanxiang Che, Xiaolong Li, and Ting Liu.
  2020.
\newblock \href {https://doi.org/10.18653/v1/2020.acl-main.10} {Slot-consistent
  {NLG} for task-oriented dialogue systems with iterative rectification
  network}.
\newblock In \emph{Proceedings of the 58th Annual Meeting of the Association
  for Computational Linguistics}, pages 97--106, Online. Association for
  Computational Linguistics.

\bibitem[{Liang et~al.(2020)Liang, Yin, Gui, Du, He, and Xu}]{liang2020aspect}
Bin Liang, Rongdi Yin, Lin Gui, Jiachen Du, Yulan He, and Ruifeng Xu. 2020.
\newblock Aspect-invariant sentiment features learning: Adversarial multi-task
  learning for aspect-based sentiment analysis.
\newblock In \emph{Proceedings of the 29th ACM International Conference on
  Information \& Knowledge Management}, pages 825--834.

\bibitem[{Liu et~al.(2019)Liu, Ott, Goyal, Du, Joshi, Chen, Levy, Lewis,
  Zettlemoyer, and Stoyanov}]{liu2019roberta}
Yinhan Liu, Myle Ott, Naman Goyal, Jingfei Du, Mandar Joshi, Danqi Chen, Omer
  Levy, Mike Lewis, Luke Zettlemoyer, and Veselin Stoyanov. 2019.
\newblock \href {http://arxiv.org/abs/1907.11692} {Roberta: A robustly
  optimized bert pretraining approach}.

\bibitem[{Madotto et~al.(2018)Madotto, Wu, and
  Fung}]{madotto-etal-2018-mem2seq}
Andrea Madotto, Chien-Sheng Wu, and Pascale Fung. 2018.
\newblock \href {https://doi.org/10.18653/v1/P18-1136} {{M}em2{S}eq:
  Effectively incorporating knowledge bases into end-to-end task-oriented
  dialog systems}.
\newblock In \emph{Proceedings of the 56th Annual Meeting of the Association
  for Computational Linguistics (Volume 1: Long Papers)}, pages 1468--1478,
  Melbourne, Australia. Association for Computational Linguistics.

\bibitem[{Nie et~al.(2021)Nie, Williamson, Bansal, Kiela, and
  Weston}]{nie-etal-2021-like}
Yixin Nie, Mary Williamson, Mohit Bansal, Douwe Kiela, and Jason Weston. 2021.
\newblock \href {https://doi.org/10.18653/v1/2021.acl-long.134} {{I} like fish,
  especially dolphins: Addressing contradictions in dialogue modeling}.
\newblock In \emph{Proceedings of the 59th Annual Meeting of the Association
  for Computational Linguistics and the 11th International Joint Conference on
  Natural Language Processing (Volume 1: Long Papers)}, pages 1699--1713,
  Online. Association for Computational Linguistics.

\bibitem[{Peng et~al.(2020)Peng, Zhu, Li, Li, Li, Zeng, and
  Gao}]{peng-etal-2020-shot}
Baolin Peng, Chenguang Zhu, Chunyuan Li, Xiujun Li, Jinchao Li, Michael Zeng,
  and Jianfeng Gao. 2020.
\newblock \href {https://doi.org/10.18653/v1/2020.findings-emnlp.17} {Few-shot
  natural language generation for task-oriented dialog}.
\newblock In \emph{Findings of the Association for Computational Linguistics:
  EMNLP 2020}, pages 172--182, Online. Association for Computational
  Linguistics.

\bibitem[{Qin et~al.(2020{\natexlab{a}})Qin, Che, Li, Ni, and
  Liu}]{Qin_Che_Li_Ni_Liu_2020}
Libo Qin, Wanxiang Che, Yangming Li, Mingheng Ni, and Ting Liu.
  2020{\natexlab{a}}.
\newblock \href {https://doi.org/10.1609/aaai.v34i05.6391} {Dcr-net: A deep
  co-interactive relation network for joint dialog act recognition and
  sentiment classification}.
\newblock \emph{Proceedings of the AAAI Conference on Artificial Intelligence},
  34(05):8665--8672.

\bibitem[{Qin et~al.(2019{\natexlab{a}})Qin, Che, Li, Wen, and
  Liu}]{qin-etal-2019-stack}
Libo Qin, Wanxiang Che, Yangming Li, Haoyang Wen, and Ting Liu.
  2019{\natexlab{a}}.
\newblock \href {https://doi.org/10.18653/v1/D19-1214} {A stack-propagation
  framework with token-level intent detection for spoken language
  understanding}.
\newblock In \emph{Proceedings of the 2019 Conference on Empirical Methods in
  Natural Language Processing and the 9th International Joint Conference on
  Natural Language Processing (EMNLP-IJCNLP)}, pages 2078--2087, Hong Kong,
  China. Association for Computational Linguistics.

\bibitem[{Qin et~al.(2021)Qin, Liu, Che, Kang, Zhao, and
  Liu}]{qin2021cointeractive}
Libo Qin, Tailu Liu, Wanxiang Che, Bingbing Kang, Sendong Zhao, and Ting Liu.
  2021.
\newblock \href {http://arxiv.org/abs/2010.03880} {A co-interactive transformer
  for joint slot filling and intent detection}.

\bibitem[{Qin et~al.(2019{\natexlab{b}})Qin, Liu, Che, Wen, Li, and
  Liu}]{qin-etal-2019-entity}
Libo Qin, Yijia Liu, Wanxiang Che, Haoyang Wen, Yangming Li, and Ting Liu.
  2019{\natexlab{b}}.
\newblock \href {https://doi.org/10.18653/v1/D19-1013} {Entity-consistent
  end-to-end task-oriented dialogue system with {KB} retriever}.
\newblock In \emph{Proceedings of the 2019 Conference on Empirical Methods in
  Natural Language Processing and the 9th International Joint Conference on
  Natural Language Processing (EMNLP-IJCNLP)}, pages 133--142, Hong Kong,
  China. Association for Computational Linguistics.

\bibitem[{Qin et~al.(2020{\natexlab{b}})Qin, Xu, Che, Zhang, and
  Liu}]{qin-etal-2020-dynamic}
Libo Qin, Xiao Xu, Wanxiang Che, Yue Zhang, and Ting Liu. 2020{\natexlab{b}}.
\newblock \href {https://doi.org/10.18653/v1/2020.acl-main.565} {Dynamic fusion
  network for multi-domain end-to-end task-oriented dialog}.
\newblock In \emph{Proceedings of the 58th Annual Meeting of the Association
  for Computational Linguistics}, pages 6344--6354, Online. Association for
  Computational Linguistics.

\bibitem[{Song et~al.(2020)Song, Wang, Zhang, Zhao, Liu, and
  Liu}]{song-etal-2020-profile}
Haoyu Song, Yan Wang, Wei-Nan Zhang, Zhengyu Zhao, Ting Liu, and Xiaojiang Liu.
  2020.
\newblock \href {https://doi.org/10.18653/v1/2020.emnlp-main.539} {Profile
  consistency identification for open-domain dialogue agents}.
\newblock In \emph{Proceedings of the 2020 Conference on Empirical Methods in
  Natural Language Processing (EMNLP)}, pages 6651--6662, Online. Association
  for Computational Linguistics.

\bibitem[{Takanobu et~al.(2020)Takanobu, Liang, and
  Huang}]{takanobu-etal-2020-multi}
Ryuichi Takanobu, Runze Liang, and Minlie Huang. 2020.
\newblock \href {https://doi.org/10.18653/v1/2020.acl-main.59} {Multi-agent
  task-oriented dialog policy learning with role-aware reward decomposition}.
\newblock In \emph{Proceedings of the 58th Annual Meeting of the Association
  for Computational Linguistics}, pages 625--638, Online. Association for
  Computational Linguistics.

\bibitem[{Talmor et~al.(2019)Talmor, Herzig, Lourie, and
  Berant}]{talmor-etal-2019-commonsenseqa}
Alon Talmor, Jonathan Herzig, Nicholas Lourie, and Jonathan Berant. 2019.
\newblock \href {https://doi.org/10.18653/v1/N19-1421} {{C}ommonsense{QA}: A
  question answering challenge targeting commonsense knowledge}.
\newblock In \emph{Proceedings of the 2019 Conference of the North {A}merican
  Chapter of the Association for Computational Linguistics: Human Language
  Technologies, Volume 1 (Long and Short Papers)}, pages 4149--4158,
  Minneapolis, Minnesota. Association for Computational Linguistics.

\bibitem[{Wang et~al.(2020)Wang, Liu, Bi, Liu, He, Xu, and
  Yang}]{wang-etal-2020-dual}
Jian Wang, Junhao Liu, Wei Bi, Xiaojiang Liu, Kejing He, Ruifeng Xu, and Min
  Yang. 2020.
\newblock \href {https://doi.org/10.18653/v1/2020.coling-main.362} {Dual
  dynamic memory network for end-to-end multi-turn task-oriented dialog
  systems}.
\newblock In \emph{Proceedings of the 28th International Conference on
  Computational Linguistics}, pages 4100--4110, Barcelona, Spain (Online).
  International Committee on Computational Linguistics.

\bibitem[{Welleck et~al.(2019)Welleck, Weston, Szlam, and
  Cho}]{welleck-etal-2019-dialogue}
Sean Welleck, Jason Weston, Arthur Szlam, and Kyunghyun Cho. 2019.
\newblock \href {https://doi.org/10.18653/v1/P19-1363} {Dialogue natural
  language inference}.
\newblock In \emph{Proceedings of the 57th Annual Meeting of the Association
  for Computational Linguistics}, pages 3731--3741, Florence, Italy.
  Association for Computational Linguistics.

\bibitem[{Wen et~al.(2018)Wen, Liu, Che, Qin, and Liu}]{wen-etal-2018-sequence}
Haoyang Wen, Yijia Liu, Wanxiang Che, Libo Qin, and Ting Liu. 2018.
\newblock \href {https://aclanthology.org/C18-1320} {Sequence-to-sequence
  learning for task-oriented dialogue with dialogue state representation}.
\newblock In \emph{Proceedings of the 27th International Conference on
  Computational Linguistics}, pages 3781--3792, Santa Fe, New Mexico, USA.
  Association for Computational Linguistics.

\bibitem[{Wu et~al.(2019{\natexlab{a}})Wu, Madotto, Hosseini-Asl, Xiong,
  Socher, and Fung}]{wu-etal-2019-transferable}
Chien-Sheng Wu, Andrea Madotto, Ehsan Hosseini-Asl, Caiming Xiong, Richard
  Socher, and Pascale Fung. 2019{\natexlab{a}}.
\newblock \href {https://doi.org/10.18653/v1/P19-1078} {Transferable
  multi-domain state generator for task-oriented dialogue systems}.
\newblock In \emph{Proceedings of the 57th Annual Meeting of the Association
  for Computational Linguistics}, pages 808--819, Florence, Italy. Association
  for Computational Linguistics.

\bibitem[{Wu et~al.(2019{\natexlab{b}})Wu, Socher, and
  Xiong}]{wu2019globaltolocal}
Chien-Sheng Wu, Richard Socher, and Caiming Xiong. 2019{\natexlab{b}}.
\newblock \href {http://arxiv.org/abs/1901.04713} {Global-to-local memory
  pointer networks for task-oriented dialogue}.

\bibitem[{Xu et~al.(2021)Xu, Li, Zhu, Zhang, and Zhou}]{xu-etal-2021-xlpt}
Dongqin Xu, Junhui Li, Muhua Zhu, Min Zhang, and Guodong Zhou. 2021.
\newblock \href {https://doi.org/10.18653/v1/2021.acl-long.73} {{XLPT}-{AMR}:
  Cross-lingual pre-training via multi-task learning for zero-shot {AMR}
  parsing and text generation}.
\newblock In \emph{Proceedings of the 59th Annual Meeting of the Association
  for Computational Linguistics and the 11th International Joint Conference on
  Natural Language Processing (Volume 1: Long Papers)}, pages 896--907, Online.
  Association for Computational Linguistics.

\bibitem[{Yang et~al.(2019)Yang, Zhang, Gao, Ji, and
  Chen}]{yang-etal-2019-simple}
Runqi Yang, Jianhai Zhang, Xing Gao, Feng Ji, and Haiqing Chen. 2019.
\newblock \href {https://doi.org/10.18653/v1/P19-1465} {Simple and effective
  text matching with richer alignment features}.
\newblock In \emph{Proceedings of the 57th Annual Meeting of the Association
  for Computational Linguistics}, pages 4699--4709, Florence, Italy.
  Association for Computational Linguistics.

\bibitem[{Yang et~al.(2020)Yang, Dai, Yang, Carbonell, Salakhutdinov, and
  Le}]{yang2020xlnet}
Zhilin Yang, Zihang Dai, Yiming Yang, Jaime Carbonell, Ruslan Salakhutdinov,
  and Quoc~V. Le. 2020.
\newblock \href {http://arxiv.org/abs/1906.08237} {Xlnet: Generalized
  autoregressive pretraining for language understanding}.

\bibitem[{Young et~al.(2013)Young, Gasic, Thomson, and
  Williams}]{DBLP:journals/pieee/YoungGTW13}
Steve~J. Young, Milica Gasic, Blaise Thomson, and Jason~D. Williams. 2013.
\newblock \href {https://doi.org/10.1109/JPROC.2012.2225812} {Pomdp-based
  statistical spoken dialog systems: {A} review}.
\newblock \emph{Proceedings of the {IEEE}}, 101(5):1160--1179.

\bibitem[{Zhang et~al.(2018)Zhang, Dinan, Urbanek, Szlam, Kiela, and
  Weston}]{zhang-etal-2018-personalizing}
Saizheng Zhang, Emily Dinan, Jack Urbanek, Arthur Szlam, Douwe Kiela, and Jason
  Weston. 2018.
\newblock \href {https://doi.org/10.18653/v1/P18-1205} {Personalizing dialogue
  agents: {I} have a dog, do you have pets too?}
\newblock In \emph{Proceedings of the 56th Annual Meeting of the Association
  for Computational Linguistics (Volume 1: Long Papers)}, pages 2204--2213,
  Melbourne, Australia. Association for Computational Linguistics.

\bibitem[{Zheng et~al.(2020)Zheng, Chen, Huang, Liu, and
  Zhu}]{zheng2020personalized}
Yinhe Zheng, Guanyi Chen, Minlie Huang, Song Liu, and Xuan Zhu. 2020.
\newblock \href {http://arxiv.org/abs/1901.09672} {Personalized dialogue
  generation with diversified traits}.

\end{thebibliography}
\bibliographystyle{acl_natbib}


\end{document}